\documentclass[conference]{IEEEtran}
\IEEEoverridecommandlockouts
\usepackage{cite}
\usepackage{amsmath,amssymb,amsfonts}
\usepackage{algorithmic}
\usepackage{graphicx}
\usepackage{textcomp}
\usepackage{xcolor}
\def\BibTeX{{\rm B\kern-.05em{\sc i\kern-.025em b}\kern-.08em
    T\kern-.1667em\lower.7ex\hbox{E}\kern-.125emX}}
\begin{document}

\title{Gesture Recognition with mmWave Wi-Fi Access Points: Lessons Learned\\

}

\author{\IEEEauthorblockN{Nabeel Nisar Bhat}
\IEEEauthorblockA{\textit{IDLab-Faculty of Science}\\
\textit{University of Antwerp-imec}\\
Antwerp, Belgium\\
nabeelnisar.bhat@uantwerpen.be
}
\and
\IEEEauthorblockN{Rafael Berkvens}
\IEEEauthorblockA{\textit{IDLab-Faculty of Applied Engineering}\\
\textit{University of Antwerp-imec}\\
Antwerp, Belgium\\
rafael.berkvens@uantwerpen.be
}
\and
\IEEEauthorblockN{Jeroen Famaey}
\IEEEauthorblockA{\textit{IDLab-Faculty of Science}\\
\textit{University of Antwerp-imec}\\
Antwerp, Belgium\\
jeroen.famaey@uantwerpen.be
}}

\maketitle 

\begin{abstract}
In recent years, channel state information (CSI) at sub-6 GHz has been widely exploited for Wi-Fi sensing, particularly for activity and gesture recognition. In this work, we instead explore mmWave (60 GHz) Wi-Fi signals for gesture recognition/pose estimation. Our focus is on the mmWave Wi-Fi signals so that they can be used not only for high data rate communication but also for improved sensing \textit{e.g.}, for extended reality (XR) applications. For this reason, we extract spatial beam signal-to-noise ratios (SNRs) from the periodic beam training employed by IEEE 802.11ad devices. We consider a set of 10 gestures/poses motivated by XR applications. We conduct experiments in two environments and with three people. As a comparison, we also collect CSI from IEEE 802.11ac devices. To extract features from the CSI and the beam SNR, we leverage a deep neural network (DNN). The DNN classifier achieves promising results on the beam SNR task with state-of-the-art 96.7\% accuracy in a single environment, even with a limited dataset. We also investigate the robustness of the beam SNR against CSI across different environments. Our experiments reveal that features from the CSI generalize without additional re-training, while those from beam SNRs do not. Therefore, re-training is required in the latter case. 
\end{abstract}

\begin{IEEEkeywords}
Wi-Fi signals, context aware, human activity recognition, gesture recognition, millimeter-wave, CSI, beam SNR, deep neural networks.
\end{IEEEkeywords}

\section{Introduction}
In the past few years, wireless signals have caught significant attention, and a new field known as wireless sensing has emerged. In the wireless sensing world, most of the focus has been on Wi-Fi signals and, consequently, on Wi-Fi-sensing, primarily due to the availability of such devices. Wi-Fi signals at 2.4 and 5 GHz have been widely used for activity recognition \cite{wang2014eyes,wang2015understanding,palipana2018falldefi,meneghello2022sharp}, gesture recognition/pose estimation \cite{abdelnasser2015wigest,he2015wig,tan2016wifinger,ma2018signfi,zheng2019zero,wang2021point,ren2022gopose},
human detection \cite{gong2016adaptive,qian2018enabling,hang2019wish}, localization \cite{wu2012csi,foliadis2021csi,wang2016csi}, crowd counting \cite{depatla2018crowd}, biological activities \cite{abdelnasser2015ubibreathe,gu2019wifi}, and human identification \cite{zhang2016wifi,zhang2020gate,wang2019csiid}. Most of these works exploit physical layer parameters, such as channel state information (CSI), from Wi-Fi devices and use a machine learning/deep learning approach to classify such activities. Compared to camera-based activity recognition, Wi-Fi has improved privacy, works in non-line-of-sight, and does not require a well-lit environment \cite{smith2018gesture}. Moreover, Wi-Fi signals contain significant information about the environment. Due to their low cost, Wi-Fi devices suffer from hardware impairments such as sampling frequency offset (SFO), carrier frequency offset (CFO), multi-path propagation, and fading. Therefore, environment robustness/generalization is one of the key challenges with Wi-Fi sensing.
\par Among the Wi-Fi standards, much of the focus has been on 2.4 GHz and 5 GHz frequency bands (\textit{e.g.}, IEEE 802.11n and IEEE 802.11ac). However, radio signals at these frequencies have limited resolution due to low bandwidth. Recently, the focus has shifted to mmWave signals ($>$30 GHz), which offer several advantages compared to sub-6 GHz. Apart from increased data rates for communication, mmWaves have a dominant line-of-sight component with respect to non-line-of-sight \cite{peinecke2008phong}. Also, due to larger bandwidth, fine delay resolution of multi-path components can be achieved \cite{de2021convergent}. Thanks to the narrow-beam or pencil-beam \cite{de2021convergent}, the spatial resolution is very high at mmWave frequencies. 
Therefore, by moving to mmWave, improved sensing is possible. mmWave radars have been widely used for gesture recognition in human-computer interactions \cite{smith2018gesture},\cite{lien2016soli} \cite{ren2021hand}. However, in this work, we exploit commercial-off-the-shelf (COTS) mmWave Wi-Fi as opposed to dedicated radar. Our focus is on joint communication and sensing (JCAS), in the sense that we want to use the same signals for communication as well as for sensing. Using the same signals for communication and sensing gives us several advantages, such as cost and availability.
\par In this work, we demonstrate gesture/pose recognition capabilities of mmWave Wi-Fi devices, which have not yet been explored much. We extract beam signal-to-noise ratios (SNRs) from the periodic sector sweep algorithm employed by IEEE 802.11ad devices. Compared to CSI, beam SNRs can be considered a direct indicator of channel quality \cite{yu2020human}. Based on the beam SNRs, we train a deep neural network (DNN) to extract features and map changes in SNRs to different gestures. We perform multiple experiments with different people, different environments, and different orientations of people. Our gestures are motivated by the extended reality (XR) applications. Since XR applications require high data rates, which can be achieved using mmWave access points\cite{perfecto2019mobile}, our focus is on the sensing capabilities of these devices so that device-free (hands-free) sensing can be achieved in the XR world. Therefore, we investigate the performance of mid-grained beam SNRs and compare it to fine-grained CSI captured from a 5 GHz Wi-Fi access point with an application for gesture recognition. The scope of this research is not to develop new deep learning pipelines for CSI/beam SNR-based gesture recognition. Instead, we use existing convolutional neural network (CNN) architectures and tailor them according to the data and needs to demonstrate the performance of 60 GHz beam SNR. Our experiments with 10 XR-related gestures in natural environments reveal exciting results. Though gesture recognition can be performed reliably even with low-sample beam SNR, we learn some interesting lessons from our trials for future improvement.
\par The rest of the paper is organized as follows. First, we discuss the related work in Section \ref{related work} concerning CSI and beam SNR. The related work highlights recent developments in Wi-Fi sensing at sub-6 GHz and 60 GHz frequencies. Then in Section \ref{data}, we describe the hardware setup and the details of the gestures/poses. This is followed by Section \ref{meth}, where we describe our method, \textit{i.e.,} deep neural network, to extract the features from the data. Finally, we describe the experiments and performance of our method in Section \ref{exp}.
\section{Related Work} \label{related work}
Activity recognition is the process of identifying the actions of a subject (\textit{e.g.,} human or robot) from a series of observations. It involves detecting movements such as fall, sitting, standing, and running. On the other hand, gesture recognition involves subtle movements like lifting a hand, swiping, and arm movements. The latter is much more challenging than coarse activity recognition. This section reviews the existing literature on gesture recognition based on Wi-Fi. Our focus is on the works which exploit CSI and beam SNR at sub-6 GHz and 60 GHz, respectively. 
\subsection{Channel State Information (CSI)}

WiG \cite{he2015wig} is one of the pioneering works in Wi-Fi sensing. WiG is a low-cost device-free gesture recognition system based on CSI. The authors use a Linux 802.11n CSI tool  \cite{halperin2011tool} to extract CSI from Intel 5300 network interface card (NIC). Wavelet denoising is used to clean and smoothen the raw CSI data. After cleaning CSI, outliers are removed, and finally, data is fed to the Support Vector Machine (SVM) classifier. The method achieves $92\%$ accuracy in classifying four different gestures under line-of-sight conditions. Abdulaziz et al.\cite{al2016wiger} describe a device-free gesture recognition system based on CSI. It uses dynamic time warping to classify different hand gestures. Instead of classification, recent works exploit Wi-Fi signals to construct 2D human poses for fine-grained perception. For example, Wang et al. \cite{wang2019person} use RGB camera images as annotations. Using deep learning, 2D body pose coordinates are reconstructed from Wi-Fi signals. The results obtained are quite similar to computer vision based methods on 2D images. However, the study is conducted in a single environment.  
\par Since CSI varies under different environments and the fact that different people perform gestures differently, the focus has recently shifted to robustness and generalization of CSI-based gesture recognition, in addition to multi-people sensing. Jian et al.\cite{jiang2018towards} present El, a deep-learning-based approach to achieve domain-independent activity recognition. It incorporates an adversarial network consisting of a CNN-based feature extractor. The proposed network can extract common features among different domains. Similarly, the authors propose CrossSense\cite{zhang2018crosssense}, an artificial neural network, to generate virtual data for the new environment from previously collected measurements. The authors use transfer learning to train a network with fewer data from the unseen site. Transfer learning reduces the training cost in unseen environments. Differently, the authors propose a cross-domain gesture-recognition system \cite{zheng2019zero} that involves extracting a domain-independent feature called body-coordinate velocity profile (BVP) from the gestures. With velocity profiles, the authors develop a model trained only once and can adapt to unseen domains and orientations without re-training. Similar to \cite{zheng2019zero}, Jiang et al. propose WiPose \cite{jiang2020towards}, the first 3D human pose construction from Wi-Fi signals and use a velocity profile for separating posture-specific features from static objects in the environment. WiPose uses a VICON system to generate 3D skeletons as ground truth. It then leverages a DNN to construct 3D skeletons from CSI at 5 GHz. WiPose achieves a 2.83 cm average error in localizing each skeletal joint. Winect \cite{ren2021winect} is another gesture recognition system that outputs a $3$D skeleton of a human body using a $2$D angle of arrival information from reflected Wi-Fi signals. Winect is environment independent and does not rely on predefined activities. Notably, it can track free-form movements for human-computer interaction (HCI).
\par More challenging works involve extending gesture recognition to multiple people simultaneously. Venkatnarayan et al. propose WiMU \cite{venkatnarayan2018multi}, WiFi-based multi-user gesture recognition. WiMu can recognize up to 6 simultaneously performed predefined gestures with high accuracy, around 90$\%$. It can also identify the start and end times of gestures automatically. However, it can not determine which user performed which gesture. 
\par Recently, focus has shifted to mmWave signals for gesture recognition, particularly for radar-based systems. Ren et al. \cite{ren2021hand} present a hand gesture recognition prototype with $60$ GHz mmWave technology. Radar and communication waveforms are transmitted in time-division duplex (TDD). Range-Doppler information (RDI) is obtained from the Doppler radar and exploited for gesture recognition. RDI is then fed to a CNN+LSTM model, achieving $>$95$\%$ accuracy for gesture recognition. 

\subsection{Beam Signal-to-Noise ratio (SNR): 60 GHz}
Yu et al. \cite{yu2020human} exploited beam SNR at 60 GHz for the first time in Wi-Fi sensing. The authors used beam SNR for pose estimation for a single person in one environment. A neural network was used to train the classifier.
The classifier achieves 90$\%$ accuracy in identifying the pose correctly. Moreover, quantum transfer learning \cite{koike2022quantum} was used to improve the performance of the beam SNR-based gesture recognition along 7 independent sessions in the same environment. The authors consider the first four sessions as the source domain and the latter as the target domain. Since the environment evolves and Wi-Fi settings change over time, transfer learning can be used to mitigate this issue. The quantum neural networks (QNN) achieve 90$\%$ accuracy for pose recognition when the test data comes from different domains. Apart from these works, not much has been done in the context of COTS mmWave Wi-Fi sensing. 
\par With respect to the prior art, we consider a wider range of gestures motivated by XR applications. These gestures are similar (\textit{e.g.,} left swipe and right swipe) and therefore represent a challenge for deep learning pipelines to extract gesture-specific features. Moreover, we extend our experiments to multiple people and environments to draw broader conclusions about the robustness and generalization of using beam SNR. We show that even with a limited dataset and low sample rate, beam SNR-based gesture recognition can achieve a very high accuracy of around 96.7$\%$. Thanks to the deep neural network presented in Section \ref{networksss} and the careful tuning of hyper-parameters, our classifier achieves higher accuracy than the state-of-the-art. However, beam SNR-based gesture recognition falls behind CSI in terms of generalization towards multiple people and environments. 

\section{Data Collection } \label{data}
This section briefly introduces the concept of CSI and beam SNR. Moreover, we describe our experimental setup at 5 GHz and 60 GHz. Thanks to the Nexmon tool \cite{schulz2017nexmon}, we can extract fine-grained CSI from the 5 GHz device. However, the same can not be accessed from the 60 GHz router without additional overhead. Instead, we use mid-grained beam SNR at 60 GHz, which can be acquired with zero overhead from the mmWave beam training.
Finally, we describe the collected data.

\subsection{CSI}
\begin{figure} [!t]
    \centering     \includegraphics[width=8.5 cm,scale=1]{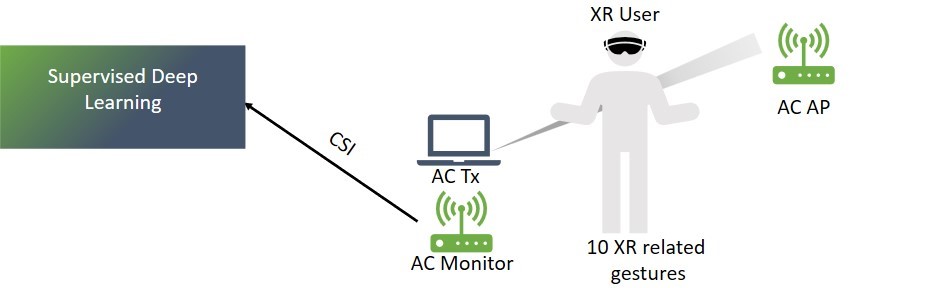}
    \caption{CSI setup at 5 GHz.}
    \label{fig:csi}
\end{figure} 

Channel state information (CSI) measures the frequency response of the channel. CSI is a measure of changes that a signal undergoes while propagating from transmitter to receiver. CSI changes based on the movement of the transmitter, receiver, or surrounding objects \cite{ma2019wifi}.
\newline Mathematically, the wireless channel can be modeled:
\begin{align} 
Y = HX + N
\end{align} where $Y$ represents the received signal,
$X$ represents the transmitted signal,
$H$ represents the CSI matrix,
and $N$ represents the noise vector.
\newline In practice, $H$ is a $4D$ tensor consisting of $N$x$M$x$K$x$T$ dimensions, where $N$ and $M$ correspond to number of transmitting and receiving antennas respectively, $K$ corresponds to number of sub-carriers, $T$ represents the packets in time. Wi-Fi devices estimate CSI with pilot symbols that are known at the transmitter. Moreover, modern Wi-Fi devices use spatial multiplexing and OFDM. Therefore, $H$ changes along space and carrier domain, on top of time domain variations. The CSI matrix is analogous to a digital image where $N$ and $M$ represent changes along space, and $K$ corresponds to color channels. Therefore, deep learning techniques used for images can be applied to CSI-based sensing tasks without much modification.
\par Our hardware setup at 5 GHz consists of two ASUS routers (RT-AC86U) and an Intel Laptop,  both supporting IEEE 802.11ac. One of the ASUS routers functions as an Access Point (AP). Since Wi-Fi devices do not expose CSI measurements to end users, we turn the other ASUS router into a CSI extractor with a firmware patch \cite{schulz2017nexmon} to access CSI measurements. We set up the system in accordance with Figure \ref{fig:csi}. So, the setup consists of two active terminals \textit{i.e.}, Intel Laptop and the ASUS AP, while the other ASUS router acts as a passive device.
The Intel transmitter and the AP exchange packets, while the patched ASUS router works as a monitor and sniffs the conversation between the AP and the laptop. A user performs gestures in the line-of-sight between the AP and the transmitter. The monitor then extracts the corresponding CSI from the packets sent back by the AP. This situation represents a real-world scenario where two Wi-Fi-enabled devices exchange packets, and a passive monitor listens to the conversation. In practice, the active terminals can be arbitrary devices exchanging Wi-Fi traffic, \textit{e.g.}, streaming XR content using the AP or accessing YouTube via Wi-Fi.
In our case, we use only one transmitting antenna ($N=1$) and one receiving antenna ($M=1$). The Intel transmitter pings 1000 packets per second of size 3008 bytes to the ASUS AP. The ASUS AP then replies with pong packets of the same size, and the corresponding CSI is captured by the monitor. The CSI is captured according to the package rate. However, the actual capture rate depends on various things, such as the CPU and memory of the device.
We collect CSI measurements at a bandwidth of 80 MHz from 256 different sub-carriers.

\subsection{Beam SNR}
Most of the research focus in COTS Wi-Fi sensing has been on sub-6 GHz frequency bands, mainly due to the existence of tools that provide access to CSI measurements for Atheros and Intel-based NICs. In contrast, at 60 GHz, few devices give access to channel measurements. The Talon AD 7200 router from TP-Link is the first IEEE 802.11ad compliant Wi-Fi device that was patched \cite{steinmetzer2017talon} to provide access to raw channel measurements in the form of beam SNR.
\begin{figure} [!t]
    \centering     \includegraphics[width=8.5 cm,scale=1]{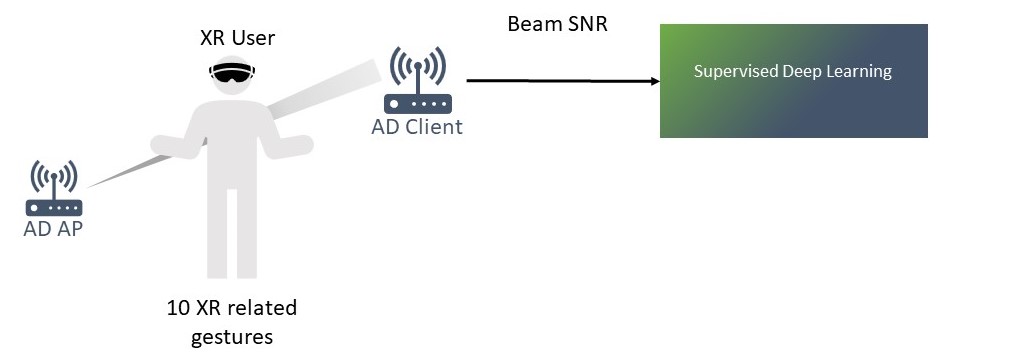}
    \caption{Beam SNR setup at 60 GHz.}
    \label{fig:beamsnr}
\end{figure} 
The router has 32 programmable antenna elements in the form of a planar array. The gain and phase of each antenna element can be controlled to change the radiation pattern of the beam. However, such a process results in huge computational costs owing to a large number of permutations and combinations ($P$). Hence, in practice, IEEE 802.11ad routers use a set of predefined patterns ($n$$<<$$P$, $n=36$) known as sectors. 802.11ad devices use a sector sweep algorithm to determine signal strength per sector, and then communication proceeds with the sector with maximum SNR. These devices periodically repeat the sector sweep to react to environmental changes or movements. The sector sweep is triggered at least once per second\cite{steinmetzer2017compressive}. In practice, the rate of sector sweep depends on the degree of obstruction along the line-of-sight. If we obstruct the line-of-sight with gestures/poses, the maximum number of sweeps per second could be as high as 10 since it is linked to the beacon interval of 102.4 milliseconds. We leave the beacon interval to the default value, this preserves the concept of JCAS as opposed to increasing the frequency, which can create an additional overhead for communication. The beam SNR is then extracted from the sector sweep frames.
The beam SNR can be modeled by the following equation:
\begin{align} \label{eq:1}
B_k =1/\sigma^2 \sum_{n=1}^{N} F_k (\theta_n) G_k{\Phi_n} = 1 	
\end{align} 
\newline where $k$ is the index of beam pattern,
$\sigma^2$ is the noise variance, $N$ is the total number of paths, $\theta_n$, and $\Phi_n$ are the azimuth angles for transmission and reception, respectively for the $n^{th}$ path.
$F_k(\theta_n)$ and $G_k(\Phi_n)$ are beam pattern gains of the transmitting and receiving antennas at the $n^{th}$ path and $k^{th}$ pattern.

\par Figure \ref{fig:beamsnr} shows experimental setup at 60 GHz. 
We use two Talon AD-7200s, one as an Access Point (AP) and the other as a client. iPerf is set between the two devices in TCP mode, resulting in a data rate of around 1Gbps. We observe that using iPerf, sector sweeps are triggered more frequently than without using any traffic exchange between the two. Moreover, it ensures that the client does not disassociate from the mmWave network during the experiment. In the absence of traffic exchange, the client sometimes disconnects, which is undesirable. Based on the toolset \cite{steinmetzer2017talon}, beam SNR measurements from $36$ sectors can be extracted from the AP and saved on a local machine. Compared to CSI, beam SNR has a relatively low sample rate which depends on the rate at which the AP performs sector sweep, as mentioned above.  \par Although the bandwidth is higher at 60 GHz, the lower sample rate and lower number of features in beam SNR are expected to lead to reduced performance of the deep learning algorithms compared to CSI-based gesture recognition.
\subsection{Details of Gestures/Poses} \label{details}
\begin{figure} [!htp]
    \centering
    \includegraphics[width=8.5 cm,scale=1]{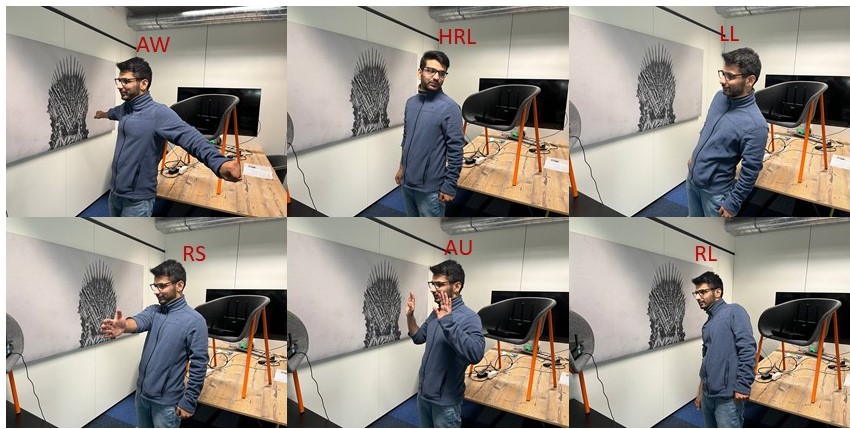}
    \caption{XR related gestures/poses}
    \label{fig:gestures}
\end{figure}
We take inspiration from XR applications and perform $10$ gestures/poses in the line-of-sight between the AP and the client. The duration of each gesture/pose is set to $15$ seconds. We define poses as fixed gestures where the person remains in a particular position for the entire $15$ seconds, while the gestures involve continuous movements for the entire duration. We collect the following poses. For the sake of brevity and space constraints, only some of them are shown in Figure \ref{fig:gestures}:
\begin{enumerate}
\item Empty (E) pose means that a person stands between the routers and does nothing. This pose serves as a baseline for other gestures/poses.
\item Right Lean (RL) involves leaning to the right.
\item Left Lean (LL) is the counterpart of the RL and therefore involves leaning to the left. 
\item Arms Up (AU) is lifting one's arms.
\item Arms Wide (AW) involves opening arms wide horizontally. 
\end{enumerate}
\par In addition, we collect the following gestures:
\begin{enumerate}
 \item Push (P) involves moving the hand forward. 
 \item Right Swipe (RS) involves swiping the hand to the right. 
 \item Left Swipe (LS) is the counterpart of RS and involves moving the same hand in the opposite direction. 
 \item  Head Rotations Left (HRL) involves moving the head to the left.
 \item Head rotations Right (HRR) involves moving the head to the right.
 \end{enumerate}
 To the best of our knowledge, HRL and HRR are unexplored in the related work. These challenging gestures and poses represent common scenarios in different XR applications. The collected dataset is openly available to the community for research and benchmarking\footnote{[https://zenodo.org/record/7813244]}.

\section{Methodology} \label{meth}
First, we patch the firmware of the ASUS and Talon routers to get access to CSI\cite{schulz2017nexmon} and beam SNR measurements \cite{steinmetzer2017talon}, respectively. For 5 GHz, we apply Medium Access Control (MAC) filtering to capture the packets from the expected device.
The user then performs each gesture along the line-of-sight between the routers for $15$ seconds according to Figures \ref{fig:csi} and \ref{fig:beamsnr}. The CSI matrix takes the shape of $x$$\times$256, where $x$ represents the number of samples in time, and 256 represents the number of subcarriers. On the other hand, the beam SNR matrix takes the shape of $y$$\times$36, where $y$ represents the number of beam SNR samples in time and 36 the number of sectors. We use only amplitude information for the CSI and ignore the phase since our focus is on beam SNRs, so we take the most straightforward approach for CSI. We manually annotate the CSI and beam SNR data for different gestures/poses. Then the raw CSI and beam SNR data are fed to the deep neural network-based classifier. Using deep neural networks has the advantage of directly extracting features from the raw data; there is no need for additional pre-processing.
\subsection{Deep Learning}
Deep Learning has been widely used for images in solving complex tasks such as image classification, object detection, and image reconstruction. In particular, Convolutional neural networks (CNNs) have been instrumental in facial recognition, biometric authentication, and autonomous driving. CNN's have been widely used for real-world applications because they are outstanding in finding patterns within the data. In the last decade, numerous CNN architectures have been proposed. For example, AlexNet, InceptionNet, VGG, and ResNet have found success in accomplishing challenging tasks. The type of architecture to use depends on the task and objectives like cost and accuracy. Inspired by the performance of neural networks for image processing tasks, CNNs have been widely adopted for Wi-Fi sensing. Recently, 1D CNNs have been used \cite{wang2019joint} for WiFi-based activity recognition. Since CSI and beam SNR are time-series data, the $1$D convolution kernel can capture patterns along the time dimension.
Based on our analysis of public datasets like ARIL \cite{wang2019joint}, and WIAR \cite{guo2019wiar}, we found that InceptionNet (GoogLeNet) with just $2$ inception modules is sufficient to have a significant improvement both in terms of computational cost and accuracy over the proposed architectures \cite{wang2019joint},\cite{guo2019wiar}. Therefore, we use GoogLeNet with $2$ inception modules to learn the CSI and beam SNR features corresponding to different gestures.
\subsection{Network} \label{networksss}
\begin{figure} [!htp]
    \centering  
    \includegraphics[width=4 cm,scale=1]{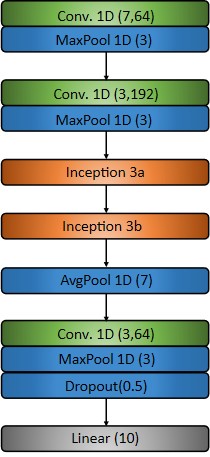}
    \caption{GoogLeNet: Depth 7}
    \label{fig:network2}
\end{figure}
GoogLeNet \cite{szegedy2015going} is a $22$ layer deep network proposed for the ImageNet Large-Scale Visual Recognition Challenge 2014 (ILSVRC2014).  
However, since our data and task are less complex, we decided to play with the architecture to find a good compromise between the computational cost and accuracy. We reduce the depth of the network to 7 layers, with only 2 inception modules. We use the same network for the beam SNR and CSI tasks. The rationale behind the inception modules is to use convolutional kernels of different sizes at the same level. As a result, a smaller kernel captures information distributed locally while a larger kernel captures information distributed globally. 
\begin{figure} [!htp]
    \centering   \includegraphics[width=8 cm,scale=1]{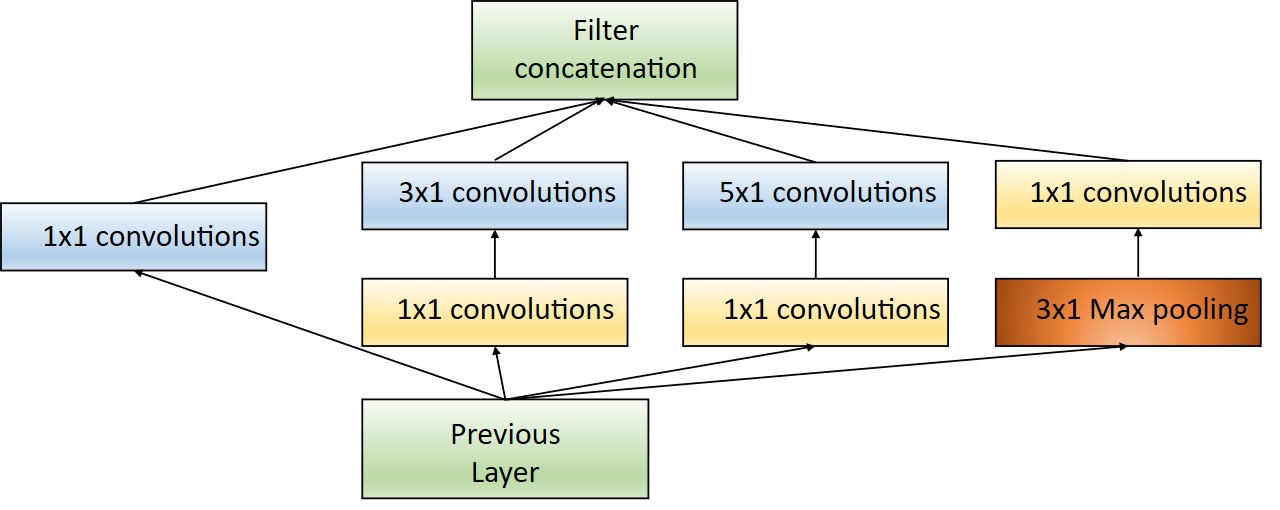}
    \caption{Inception Module}
    \label{fig:incept}
\end{figure}
Figure \ref{fig:network2} shows the architecture of the CNN-based classifier. The input is a CSI matrix/beam SNR, and accordingly, the first convolutional layer has either 256 or 36 input channels, respectively. Conv. 1D (7,64) represents the convolutional layer with kernel $7\times1$ and 64 output channels.
 The inception 3a and 3b blocks are implemented in the same way as in the original paper \cite{szegedy2015going}. After the Average Pool, the last convolutional layer reduces the number of output channels to 64. These layers extract the features from CSI and beam SNR tensors. The last layer is the fully connected layer that provides an output score for each of the 10 gestures. 
 \par Figure \ref{fig:incept} shows the inception module. The $1\times1$ convolutions (bottleneck layers) are used to decrease the computational cost by reducing the number of output channels before the input is fed to $3\times1$ and $5\times1$ convolutional layers. These convolutions don't change the size of the data. At the end, the outputs from two $1\times1$, $3\times1$, and $5\times1$ convolutions are concatenated across channels.
\section{Experiments} \label{exp}
We collect 10 gestures/poses from three humans in two different environments. The associated CSI and beam SNR variations are fed to the classifier to learn the features corresponding to these gestures.
\subsection{Testbed and experimental setup}
\begin{figure} [!htp]
    \centering
    \includegraphics[width=8.5 cm,scale=1]{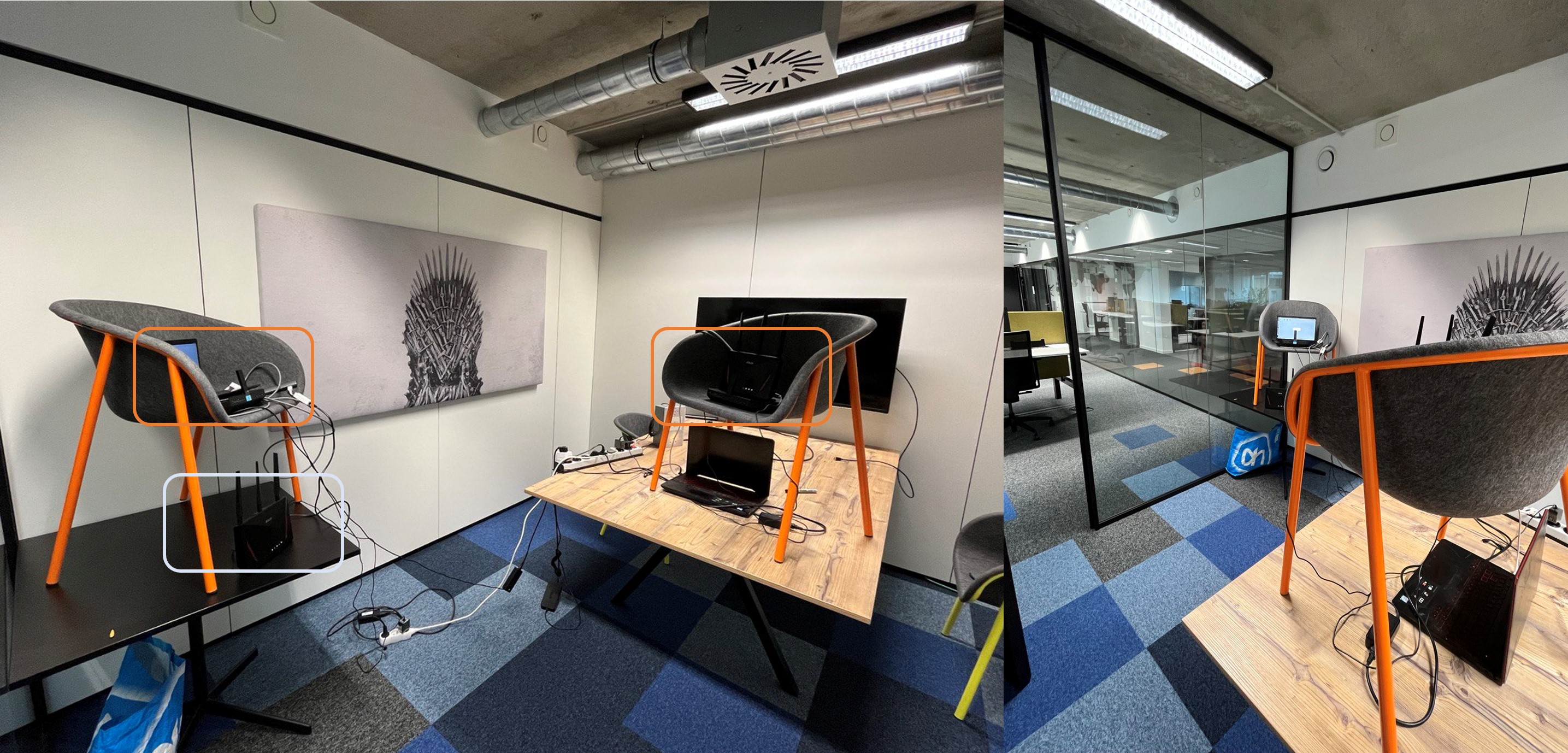}
    \caption{Testbed: Office Environment}
    \label{fig:Testbed}
\end{figure}
One of the environments is a typical living room (home), and the other is a small office space. The living room has a lot of scatterers or reflectors than the office environment. The dimensions of the home environment are 6 by 5 meters.  
While the second smaller environment measures 3.45 meters in length and 3 meters in width, for the sake of space constraints, we only show one of the environments \textit{i.e.}, office environment (cf., Figure \ref{fig:Testbed}). The room has wooden walls on three sides and a glass panel on one side with a wooden door. Therefore, the reflections are more natural. The IEEE 802.11ad AP and client and IEEE 802.11ac AP and client devices are kept on the chairs, while the IEEE 802.11ac monitor is placed below the chair. The routers are highlighted with an orange box, set 1.26 meters high with a separation of 2 meters. The light blue box indicates the position of the monitor (5 GHz). A similar setup is put in the home environment where routers are placed 1.3 meters high with a separation of 2 meters. \par We collect most of the gestures/poses in the home environment and a subset of those in the office environment. A total of 854 gesture instances for beam SNR and CSI are recorded. 

\subsection{Hyper-parameter Settings}
Hyper-parameters control the learning process of the model. They are user-defined and are used to improve the performance of the model. Tuning the deep learning model can be quite a challenging task. There are no magic rules or settings. These parameters vary from task to task and model to model. After a wide range of trial and error experiments, we use the following settings in our model. We set the number of epochs to 150 for both CSI and beam SNR tasks. We use Adam optimizer with a learning rate of 3e-4 and betas (0.9, 0.999) and epsilon values (1e-8). The batch size is set to 16 and 64 for the SNR and CSI tasks, respectively. We use the scheduler \textit{ReduceLROnPlateau} with patience of 25. The scheduler monitors the learning rate and decreases it if the metric does not improve after a patience number of epochs. This choice of parameters considerably impacts the performance of the deep learning model, especially for the beam SNR task. 

\subsection{Experiment 1: Home Environment}
The details of the gestures/poses are provided in Section \ref{details}. A single user performs gestures along the line-of-sight between the devices facing the router. Each gesture/pose lasts for 15 seconds. Each gesture is repeated around 50 times so that we have a total of 486 gesture instances of beam SNR and CSI after rejecting invalid data. For the sake of fair comparison, we use the same number of examples for the two tasks. We manually annotate the data and feed it to the deep neural network shown in Figure \ref{fig:network2} without additional processing. We split the data randomly into training and test with a standard 75:25 split, respectively. The training set consists of 364 examples, while the test data consisting of 122 gesture instances is left unseen to the classifier.
\subsection{Experiment 2: Office Environment}
Three humans perform gestures/poses along the line-of-sight between the devices. In this environment, we collected a total of 221 gestures (excluding the invalid data). We stick to the same hyper-parameter settings and the same train-test split. The training and test data consist of 165 and 56 examples, respectively. Due to the smaller dataset, we expect to see reduced performance in the office environment.

\subsection{Experiment 3: Rotated User Direction}
In the subsequent experiments, we combine the data from the two environments. We get a total of 530 and 177 training and test examples (instances), respectively. Also, we test if the deep learning model generalizes to environments or different people without re-training. \par In the above two experiments, the user faced the same direction (towards the router) while performing gestures. In an additional experiment, a single user performs gestures rotated 90$^{\circ}$ with respect to the earlier position \textit{i.e.}, we study how orientation affects the performance of the deep learning model. We collect 147 gestures with 90$^{\circ}$ rotation in the home environment. We also train on one orientation (0$^{\circ}$) and test on another (90$^{\circ}$) to see if orientation-independent gesture recognition could be achieved with CSI and beam SNR.

\subsection{Performance of the deep learning model}
\begin{figure} [!t]
    \centering     \includegraphics[width=8.5 cm,scale=1]{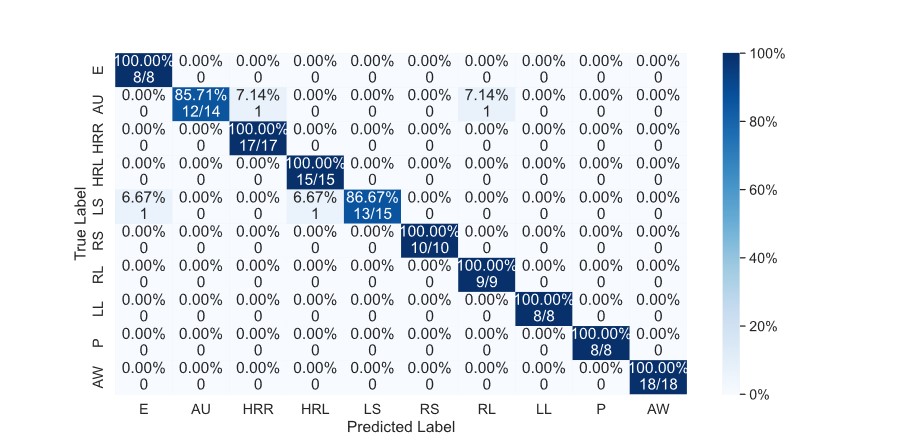}
    \caption{Performance of DNN on the beam SNR task: Home Environment.}
    \label{fig:beamsnrhome}
\end{figure}

We collect data from two environments and train the deep learning network with the training data. \begin{figure} [!h]
    \centering   \includegraphics[width=8.5 cm,scale=1]{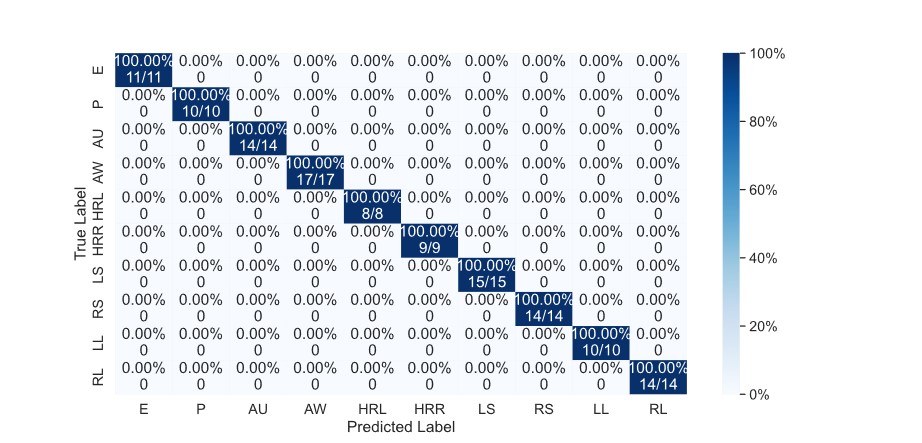}
    \caption{Performance of DNN on the CSI task: Home Environment.}
    \label{fig:confhome}
\end{figure}Then we evaluate the performance on the test data using a tool known as the confusion matrix. Figures \ref{fig:confhome} and \ref{fig:beamsnrhome} show the confusion matrices depicting the performance of the network in the home environment for the CSI and beam SNR tasks, respectively. 
With beam SNRs, we achieve a promising accuracy of around 96.7$\%$, with a per gesture accuracy above 85$\%$  even with limited data. Notably, the DNN achieves 100$\%$ accuracy with similar gestures such as RL and LL and HRR and HRL. Compared to \cite{yu2020human}, our DNN-based classifier achieves higher accuracy even with less data. On the other hand, the same DNN achieves 100$\%$ accuracy on the CSI task. With CSI, we achieve the same accuracy as \cite{schafer2021human}. However, instead of an LSTM, we use a CNN, which has a significantly lower computational cost. We believe that the gain in performance for the CSI compared to beam SNR is due to a higher number of features and a higher sample rate. This experiment validates the potential use of mid-grained low-sample beam SNRs for passive gesture recognition in challenging XR applications.
\begin{figure} [!htp]
    \centering     \includegraphics[width=8.5 cm,scale=1]{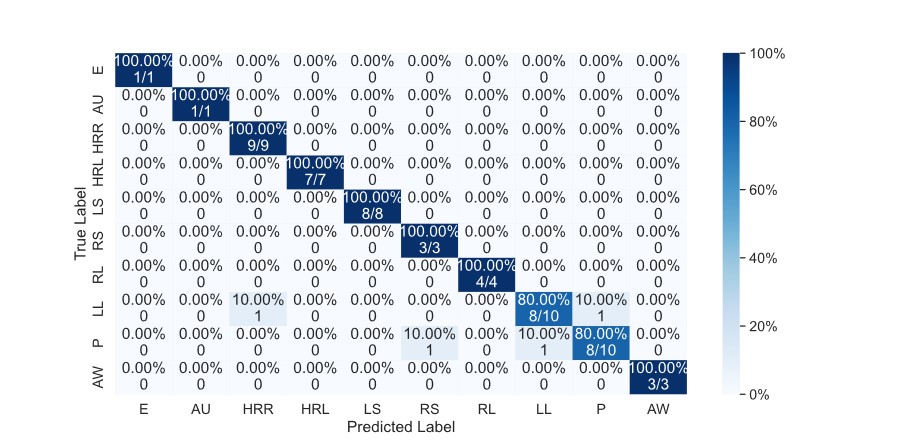}
    \caption{Performance of DNN on the beam SNR task: Office Environment.}
    \label{fig:beamsnroffice}
\end{figure}
\begin{figure} [!htp]
    \centering    \includegraphics[width=8.5 cm,scale=1]{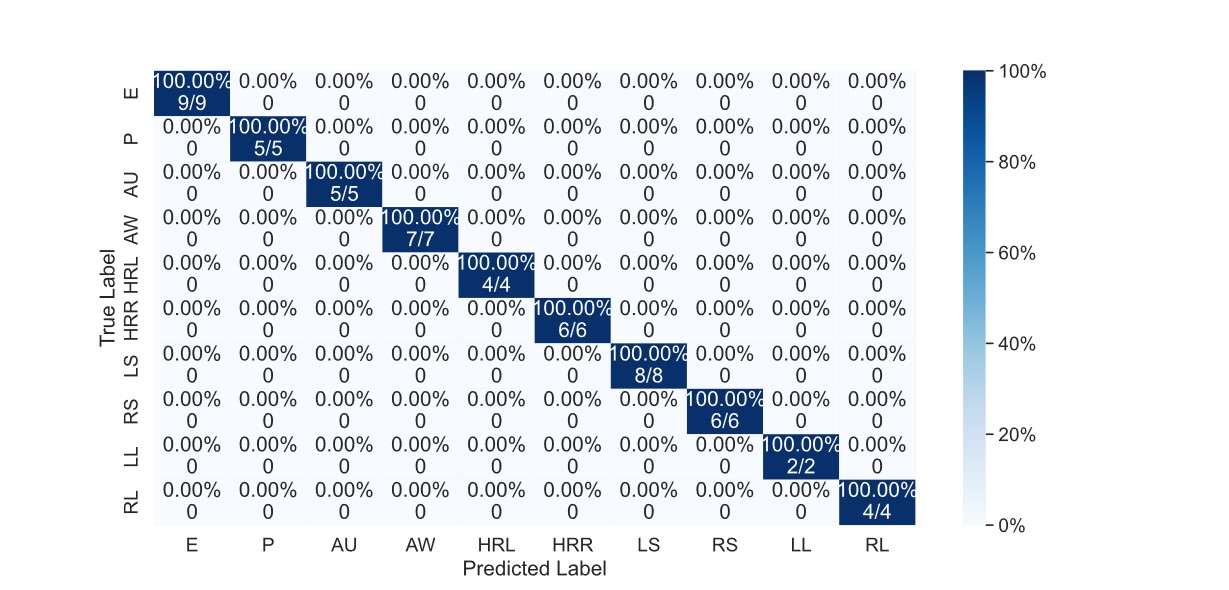}
    \caption{Performance of DNN on the CSI task: Office Environment.}
    \label{fig:CSIoffice}
\end{figure}
\par For experiment 2 in the office environment (cf., Figure \ref{fig:beamsnroffice}), where 3 users perform gestures, the beam SNR achieves an overall accuracy of around 92.8$\%$ accuracy. Also, in this case, the DNN can extract gesture-specific features from similar gestures such as HRL and HRR.
The drop in accuracy is because of the smaller dataset. While CSI still outperforms beam SNR and achieves 100$\%$ accuracy (cf., Figure \ref{fig:CSIoffice}).
\par
\begin{figure} [!htp]
    \centering    \includegraphics[width=8.5 cm,scale=1]{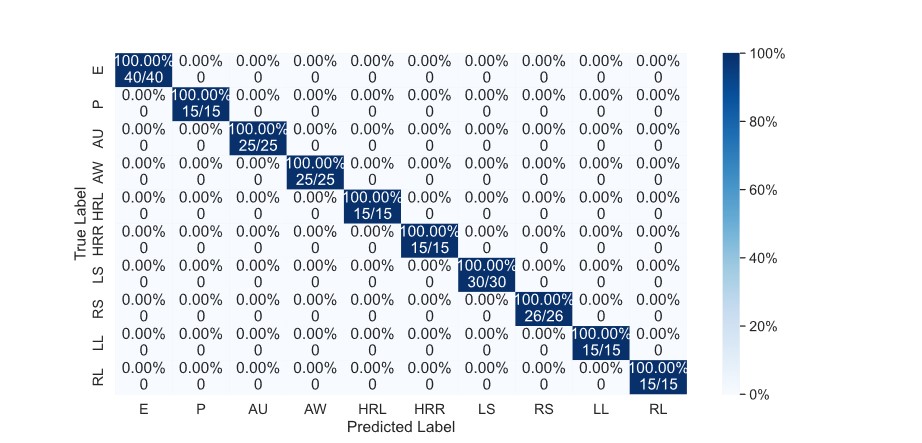}
    \caption{Generalization of DNN on the CSI task: Train Home Environment and test Office Environment.}
    \label{fig:generalization}
\end{figure}

 We then test the robustness of our neural network across multiple people and environments. We train on one environment and test on another. We see that DNN on the CSI task can generalize to different people and different environments \textit{i.e.}, a DNN trained in the home environment achieves 100$\%$ accuracy in the office environment (unseen environment) as shown in Figure \ref{fig:generalization}.
We believe that the neural network can extract deeper and more common features shared across environments from the CSI. 
\begin{figure} [!htp]
    \centering    \includegraphics[width=8.5 cm,scale=1]{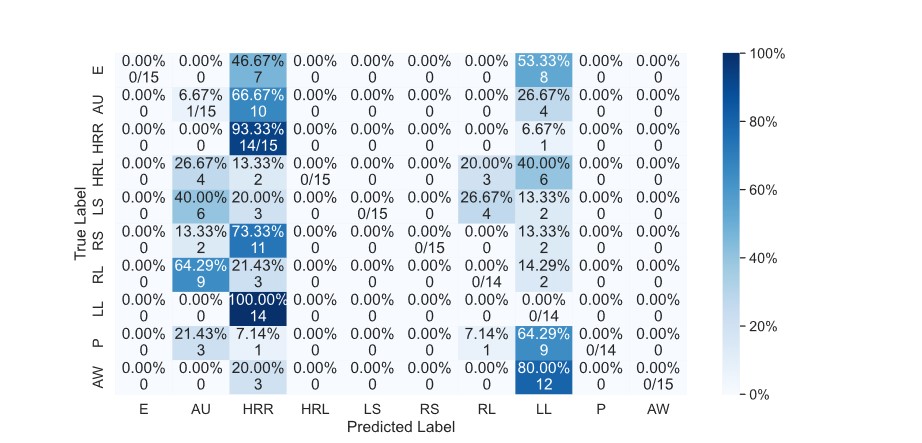}
    \caption{Generalization of DNN on the beam SNR task: Train Home Environment and test Office Environment.} \label{fig:beamsnrgeneralizationoff}
\end{figure}
\begin{figure} [!t]
    \centering    \includegraphics[width=8.5 cm,scale=1]{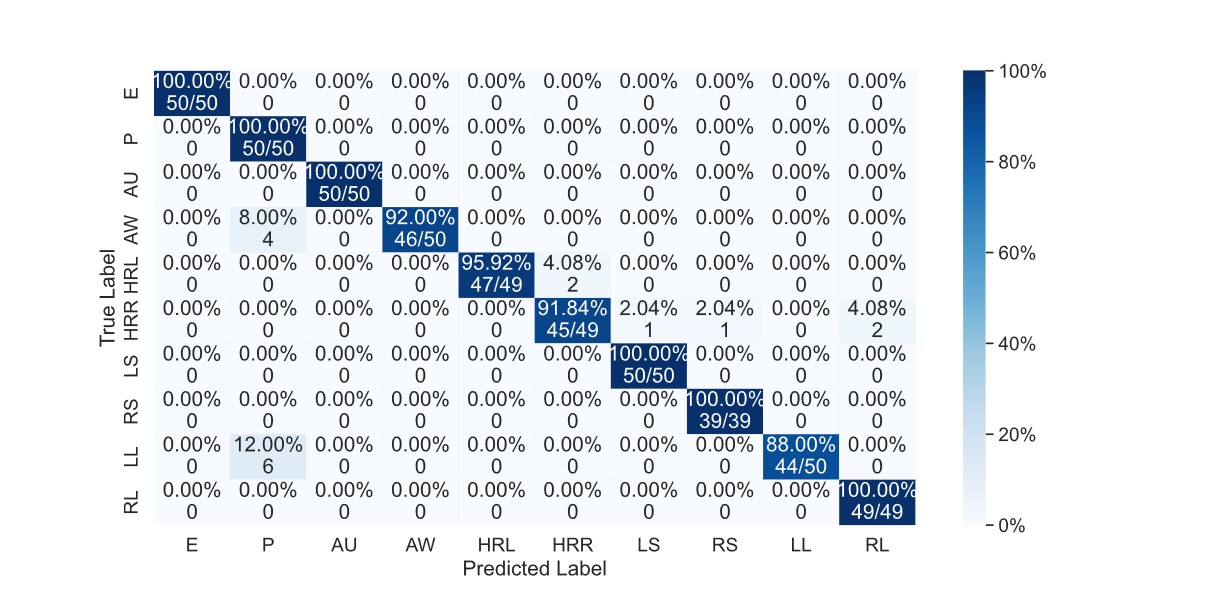}
    \caption{Generalization of DNN on the CSI task: Train Office Environment and test Home Environment.}
    \label{fig:generalizationbeacom}
\end{figure}
 While for the beam SNR, the DNN achieves 10$\%$ overall accuracy (cf., Figure \ref{fig:beamsnrgeneralizationoff}) on the unseen environment. Hence, the DNN can not generalize to different people and environments on which it was not trained. We did a reverse test too \textit{i.e.}, we trained the neural network in the office environment and tested it in the home environment. 

In this case, the DNN achieves 96.7$\%$ accuracy on the CSI task (cf., Figure \ref{fig:generalizationbeacom}). The drop in accuracy is due to a lower number of training examples in the office environment with respect to the home. Also, in this scenario, the beam SNR does not generalize across people and environments. This occurs because beam SNR is mid-grained or coarse-grained compared to fine-grained CSI. Moreover, we believe that beam SNR significantly depends on the environment compared to CSI. 
Therefore, features extracted by the DNN from CSI generalize, while those from beam SNR do not. Thus, the DNN needs the exact example of the person and environment to be able to classify gestures correctly based on the beam SNR task. Hence, for the beam SNR, re-training is required in the unseen environment. Therefore, we collected 10 additional instances per gesture per person in the second environment and fed them to the DNN for the beam SNR task.
\begin{figure} [!htp]
    \centering    \includegraphics[width=8.5 cm,scale=1]{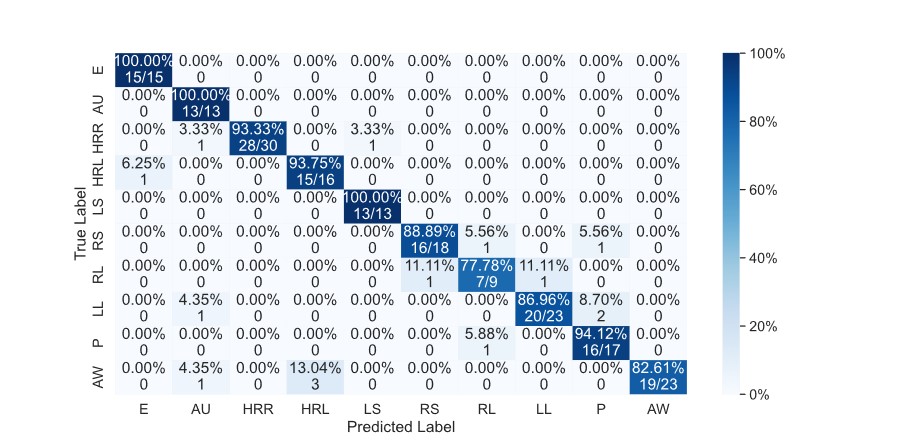}
    \caption{Re-training beam SNR with gestures from the office environment.}
    \label{fig:beamsnrmixed}
\end{figure}
Figure \ref{fig:beamsnrmixed} shows the confusion matrix for the beam SNR task across multiple (3) people and 2 environments. The DNN classifier is trained mainly using gestures from the home environment and a few additional instances from the office environment. The classifier achieves an overall accuracy of 91.5$\%$ and per gesture accuracy above 75$\%$ across two environments. On the other hand, the DNN still achieves 100$\%$ accuracy on the CSI task, which is intuitive from Figures \ref{fig:generalization} and \ref{fig:generalizationbeacom}. 
\begin{figure} [!t]
    \centering     \includegraphics[width=8.5 cm,scale=1]{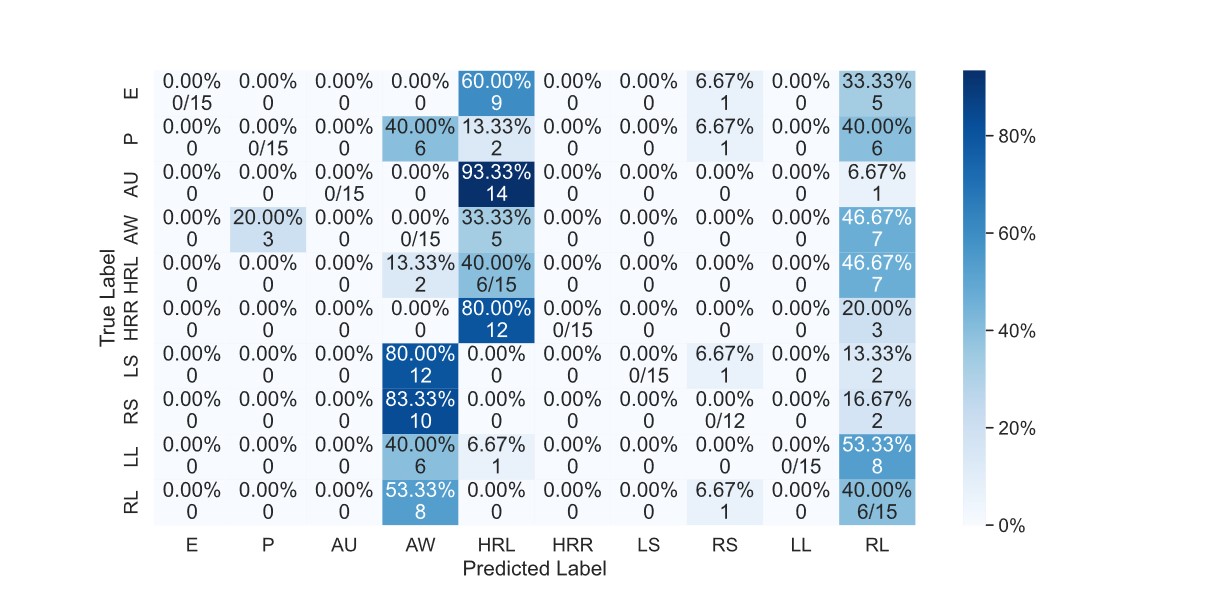}
    \caption{Generalization of DNN on the CSI task: Train 0$^{\circ}$ orientation and test 90$^{\circ}$.}
    \label{fig:orientationcsifail}
\end{figure}
\begin{figure} [!htp]
    \centering     \includegraphics[width=8.5 cm,scale=1]{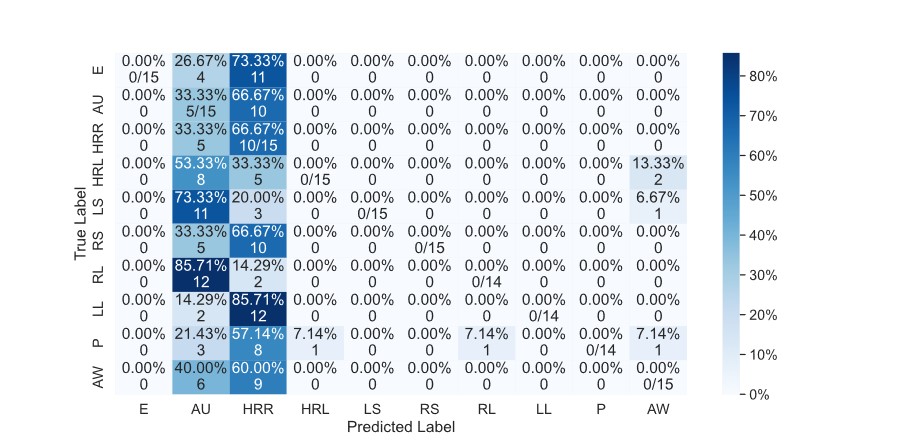}
    \caption{Generalization of DNN on the beam SNR task: Train 0$^{\circ}$ orientation and test 90$^{\circ}$.}
    \label{fig:orientationbeamsnrfail}
\end{figure}
\par Finally, we test the robustness of the DNN against the orientation of the gesture/pose. The DNN does not generalize across orientation,\textit{ i.e.,} a DNN trained in a home environment 
with gestures performed facing the router performs poorly on the dataset where gestures are performed at 90$^{\circ}$ orientation. 
The DNN achieves only 8$\%$ and 10$\%$ accuracy, as shown in Figures \ref{fig:orientationcsifail} and \ref{fig:orientationbeamsnrfail} for the CSI and beam SNR, respectively. Therefore, the DNN network needs additional re-training with the corresponding gestures in this case.  
\par
\begin{figure} [!t]
    \centering     \includegraphics[width=8.5 cm,scale=1]{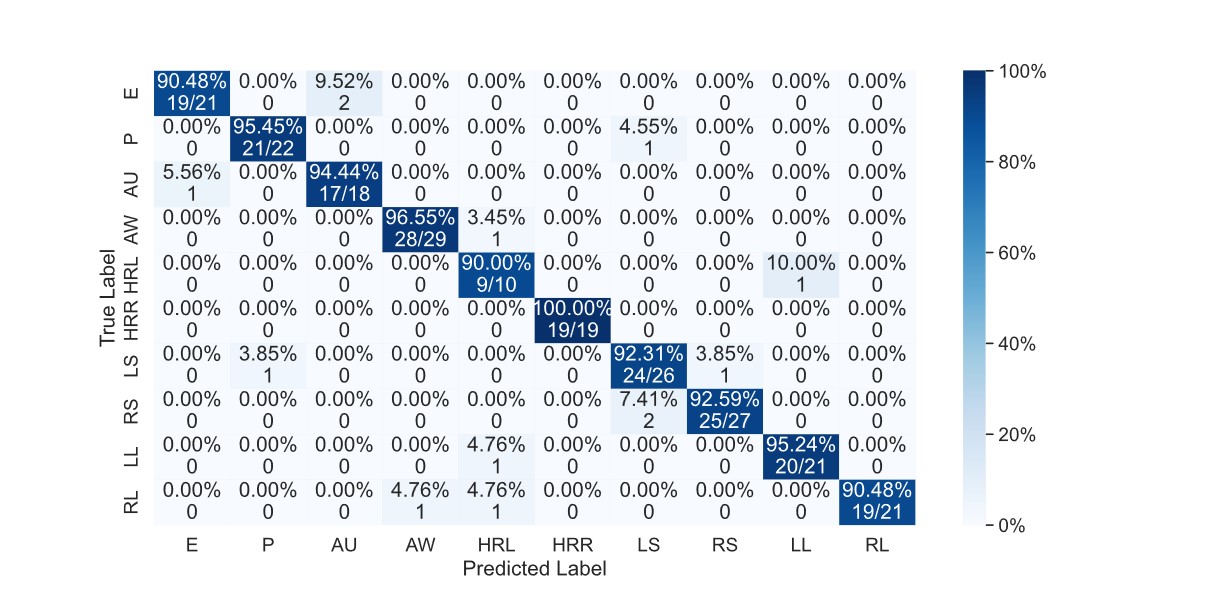}
    \caption{Performance of the DNN on two environments with the effect of orientation on the CSI task.}
    \label{fig:orientationcsi}
\end{figure}

\begin{figure} [!t]
    \centering     \includegraphics[width=8.5 cm,scale=1]{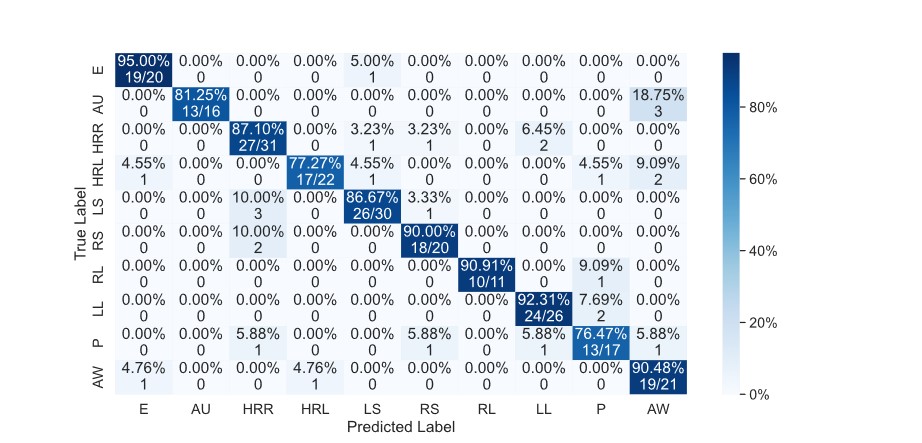}
    \caption{Performance of the DNN on two environments with the effect of orientation on the beam SNR task.}
    \label{fig:orientationbeamsnr}
\end{figure}

 Figure \ref{fig:orientationcsi} shows the
final performance of the DNN for the CSI where the training set consists of examples of gestures performed at 90$^{\circ}$ orientation in addition to 0$^{\circ}$ examples from the office and home environment (experiments 1, 2, and 3). 
The classifier achieves an overall accuracy of 93.9$\%$ against 100$\%$ in the first two experiments.  
In contrast, Figure \ref{fig:orientationbeamsnr} shows the overall performance of the DNN on the beam SNR task. The classifier achieves an overall accuracy of 87$\%$ with a per gesture accuracy above 75$\%$. In both cases, we see a drop in performance. Therefore, the orientation of the gesture impacts the performance of the DNN. This can be countered by collecting more data or working at the signal level, some pre-processing to tackle orientation. This is beyond the scope of this paper.
\section{Conclusion and Future Work} \label{Conc}
In this work, we validated the performance of the beam SNR for gesture/pose recognition, leveraging a deep neural network (DNN). We showed that even with low sample beam SNRs and a limited dataset, the DNN achieves promising results under challenging gestures relevant to XR applications in two different environments. We achieved state-of-art accuracy of around 96.7$\%$ with beam SNRs in a single environment. We also compared beam SNR with CSI and conclude that a DNN trained on CSI achieves better performance and can generalize across environments and different people. We conclude that the environment impacts the beam SNRs significantly more than the CSI. However, concerning the orientation, there is little difference between the two tasks.
Nevertheless, with minimal re-training, the DNN achieved very good results across multiple people and environments on the beam SNR task and did not lag much behind the CSI task. Concerning user orientation, the DNN performed poorly on both tasks and needs re-training with a small number of corresponding gestures. Overall, we conclude that mmWave access points can be used for Wi-Fi sensing with reliable accuracy for XR applications.  \par Our goal is to collect more data with the mmWave devices. Currently, we perform a predefined set of gestures. In our future experiments, we want to collect data with an actual Virtual Reality (VR) headset to make the gestures more natural and take any form. We will extend our data collection to more people, environments, and multi-people sensing with mmWave devices. Moreover, we also want to investigate recent explainable AI approaches to better understand how DNN is making decisions.
Currently, CNNs or LSTMs are used for Wi-Fi sensing tasks. However, they are power-hungry, representing a bottleneck when deploying such models in real-world applications. Therefore, we are also working on energy-efficient neuromorphic spiking neural networks  (SNNs) with significantly lower energy consumption than CNNs.
\section*{Acknowledgment}

This research is partly funded by the FWO WaveVR project (Grant number: G034322N).

\bibliographystyle{IEEEbib}
\bibliography{references}

\end{document}